# Adaptive Hardness-driven Augmentation and Alignment Strategies for Multi-Source Domain Adaptations

Yuxiang Yang[†], Xinyi Zeng[†], Pinxian Zeng, Chen Zu, Binyu Yan, Jiliu Zhou, *Senior Member, IEEE*, and Yan Wang*, *Member, IEEE*

*Abstract*—Multi-source Domain Adaptation (MDA) aims to transfer knowledge from multiple labeled source domains to an unlabeled target domain. Nevertheless, traditional methods primarily focus on achieving inter-domain alignment through sample-level constraints, such as Maximum Mean Discrepancy (MMD), neglecting three pivotal aspects: 1) the potential of data augmentation, 2) the significance of intra-domain alignment, and 3) the design of cluster-level constraints. In this paper, we introduce a novel hardness-driven strategy for MDA tasks, named A³MDA, which collectively considers these three aspects through **A**daptive hardness quantification and utilization in both data **A**ugmentation and domain **A**lignment. To achieve this, A³MDA progressively proposes three Adaptive Hardness Measurements (AHM), i.e., Basic, Smooth, and Comparative AHMs, each incorporating distinct mechanisms for diverse scenarios. Specifically, Basic AHM aims to gauge the instantaneous hardness for each source/target sample. Then, hardness values measured by Smooth AHM will adaptively adjust the intensity level of strong data augmentation to maintain compatibility with the model's generalization capacity. In contrast, Comparative AHM is designed to facilitate cluster-level constraints. By leveraging hardness values as sample-specific weights, the traditional MMD is enhanced into a weighted-clustered variant, strengthening the robustness and precision of inter-domain alignment. As for the often-neglected intra-domain alignment, we adaptively construct a pseudo-contrastive matrix by selecting harder samples based on the hardness rankings, enhancing the quality of pseudo-labels, and shaping a well-clustered target feature space. Experiments on multiple MDA benchmarks show that A³MDA outperforms other methods.

This work is supported by National Natural Science Foundation of China (NSFC 62371325, 62071314), Sichuan Science and Technology Program 2023YFG0263, 2023YFG0025, 2023NSFSC0497, and Opening Foundation of Agile and Intelligent Computing Key Laboratory of Sichuan Province. The associate editor coordinating the review of this manuscript and approving it for publication was ***. (* Corresponding author: Yan Wang)

† Yuxiang Yang and Xinyi Zeng have equal contributions to this work.

Yuxiang Yang, Xinyi Zeng, Pinxian Zeng, Binyu Yan, Jiliu Zhou and Yan Wang are with School of Computer Science, Sichuan University, Chengdu, China. (e-mail: yangyuxiang3@stu.scu.edu.cn; perperstudy@gmail.com; 651215874@qq.com; yanbinyu2023@163.com; zhoujiliu@cuit.edu.cn; wangyanscu@hotmail.com).

Chen Zu is with JD.com. (e-mail: chen0zu@gmail.com)

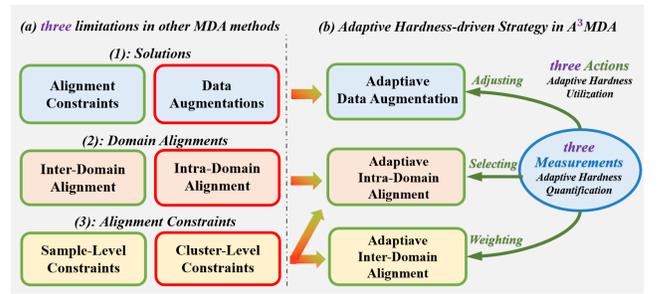

**Fig. 1.** Red boxes indicate aspects that are not handled effectively, while green boxes signify effectively handled aspects. The core of A³MDA lies in using *three*-fold **A**daptive hardness quantification and utilization to address *three* limitations.

## I. INTRODUCTION

Unsupervised Domain Adaptation (UDA) has emerged as a crucial approach for aligning distributions between labeled data (source domain) and unlabeled data (target domain). This adaptation process can be categorized into two main types, namely, Single-source Domain Adaptation (SDA) [1]-[3] and Multi-source Domain Adaptation (MDA) [4]-[6]. SDA aims to learn the underlying relationships between a single labeled source domain and an unlabeled target domain to classify the target samples accurately. However, in practical applications, labeled source data can be collected from multiple domains. Consequently, compared to SDA, MDA has garnered significant attention as a more complex and realistic scenario [7]-[10].

Despite the promising performance, most MDA studies still encounter limitations at three levels, as shown in Fig. 1(a). Firstly, prevailing approaches predominantly emphasize alignment constraints (or modules) for solutions, such as adversarial discriminator [11], correlation alignment [12], and Maximum Mean Discrepancy (MMD) [13], [14], while ignoring the potential benefits of data augmentation. This preference for alignment over augmentation is rooted in apprehensions that standard augmentation techniques in current data-driven methods [15], whether fixed or random, may *inevitably over-augment hard samples*, making them even *harder to train and potentially exceeding the model's generalization capability* [16]. Secondly, most MDA studies only address the inter-domain shift by aligning multiple source domains and a target domain into a domain-invariant feature space [5], [17]. However, they often overlook the intra-domain shifts originating from latent noise within the unlabeled target domain. *These shifts, especially noticeable in **hard target samples** near the decision boundary, can substantially undermine the categorizability of the target feature space.* In the MDA setting, such hard samples often exhibit increased uncertainty, posing challenges to a robust alignment process. Thirdly, conventional discrepancy-based methods typically employ plain constraints, such as MMD, to perform domain alignments at the sample level [18], [19]. In other words, the designed constraints indiscriminately reduce the differences between samples across domains at a coarse level, without taking into account sample-specific attributes for adaptive utilization. These attributes, including (pseudo/real)-class labels indicating their cluster memberships, and the ***hardness values*** like entropy [20]-[22] and confidence [23], [24] *that can signify their levels of learning difficulty and classification uncertainty, have the potential to enhance domain alignment with greater precision and robustness.*

Motivated to address this tripartite limitation, we propose a novel hardness-driven strategy, named $A^3MDA$, for MDA classification tasks. As shown in Fig. 1(b), all designs involving data **A**ugmentation and domain **A**lignment are **A**daptively guided by the quantification and utilization of sample-specific hardness throughout training. To achieve this, we devise three progressive Adaptive Hardness Measurements (AHM), i.e., Basic, Smooth, and Comparative AHMs. Concretely, Basic AHM is designed to gauge the difficulty of each sample and is then refined into a Smooth AHM through a historical smoothing mechanism to mitigate random fluctuations that could disrupt instantaneous hardness assessments. The hardness values from Smooth AHM are adaptively employed to adjust the intensity levels of strong data augmentations, ensuring compatibility with the model's generalization capacity. Based on Smooth AHM, Comparative AHM introduces a comparison mechanism by comparing hardness values among samples within a batch or batch-wise cluster, thereby dynamically adjusting the inter- and intra-domain alignment within cluster-level constraints. Specifically, for inter-domain alignment, the hardness values from Comparative AHM serve as sample-specific weights, in conjunction with class attributes, to enhance the traditional MMD loss into a weighted-clustered variant. This modification greatly bolsters the robustness and precision in aligning source and target distributions. As for the often-overlooked intra-domain alignment, instead of directly weighting hard samples in the loss function, we utilize their hardness rankings to adaptively select harder samples to form a pseudo-contrastive matrix. By incorporating it as an integral optimization component, $A^3MDA$ greatly improves the quality of pseudo labels and the categorizability of the target feature space. Our contributions are:

1) We introduce a novel hardness-driven strategy, $A^3MDA$, centered on a ***three***-fold adaptive hardness quantification and utilization to tackle ***three*** limitations in MDA classification tasks. Experiments on various benchmarks demonstrate its effectiveness and generalizability.

2) For adaptive hardness quantification, $A^3MDA$ employs ***three*** progressive Adaptive Hardness Modules (AHMs) - Basic, Smooth, and Comparative. Each progression incorporates unique mechanisms tailored to diverse augmentation and alignment scenarios.

3) For adaptive hardness utilization, we implement ***three*** actions: **(a)** We utilize Smooth AHM to adaptively guide the intensity of data augmentation, preventing over-augmentation and aligning with evolving generalization capabilities. **(b)** We leverage values from the Comparative AHM as sample-specific weights, transforming traditional MMD into a weighted-clustered variant to enhance the robustness and precision of inter-domain alignment. **(c)** We develop a pseudo-contrastive matrix based on selected harder samples for the often-neglected intra-domain alignment, which effectively eliminates erroneous pseudo-labels and shapes a well-clustered feature space.

## II. RELATED WORK

### A. SDA and MDA methods

Single-source Domain Adaptation (SDA) aims to transfer knowledge from a labeled source domain to an unlabeled target domain. SDA methods can be categorized into adversarial- and discrepancy-based methods. Adversarial-based methods [2], [4], [25]-[26] focus on reducing the data distribution gap between the source and target domains by employing adversarial discriminators. On the other hand, discrepancy-based methods [27]-[28] primarily utilize correlation learning or divergence learning to align source and target domains. Correlation learning focuses on directly minimizing differences in feature covariances between source and target domains [54-56], [70]. For example, Lu et al. [54] proposed a method that integrates graph embedding and sample reweighting to learn weighted correlation embeddings. In contrast, divergence losses, particularly Maximum Mean Discrepancy (MMD) and its variants, have been widely studied [13], [72]-[77] to minimize the distance between the distributions of domains. Recent advancements in SDA include innovative approaches like optimal transport [57], graph learning [71], reinforcement learning [78], and flexible strategies such as test-time adaptation [59]. Besides traditional classification tasks, SDA has also been applied to various other domains. For example, Nguyen et al. [58] proposed a cross-domain kernel classifier and applied the max-margin principle to enhance software vulnerability detection. Li et al. [74] developed a multi-source transfer learning method using optimal transport feature ranking, outperforming existing models in EEG classification.

In real-world scenarios characterized by diverse and complex source distributions, SDA methods often fall short of achieving competitive performance. Recently, Multi-source Domain Adaptation (MDA) methods have emerged to expand upon traditional SDA techniques, aiming to tackle more practical scenarios in which labeled training samples are gathered from multiple sources. For instance, STEM [29], a well-designed adversarial-guided method, employs a teacher-student architecture that provides robust theoretical guarantees regarding each component's role in domain transfer. Despite its impressive performance, STEM inevitably involves training an additional source-domain discriminator to assist the teacher expert, along with a student adversarial discriminator to minimize the gap between the mixture of source distributions and the target distribution. In contrast, MFSAN [18], a typical streamlined discrepancy-based method, relies solely on MMD without introducing additional modules for domain alignment and generates target predictions based on source-specific features directly. Despite its effectiveness, the method does not fully consider sample attributes (e.g. sample hardness and class attributes) when applying MMD constraint, as normal and hard samples are treated equally. Moreover, like most other discrepancy-based MDA classification methods, it does not address intra-domain alignment. The core idea for this issue is to enhance inter-class separability and intra-class compactness by clustering similar target samples together while pushing dissimilar samples apart. However, despite the growing focus on reducing the intra-domain gap in other research areas [60], [61], many existing unsupervised MDA methods tend to overlook this aspect, posing challenges in shaping a well-clustered feature space to achieve further improvements in the target domain.

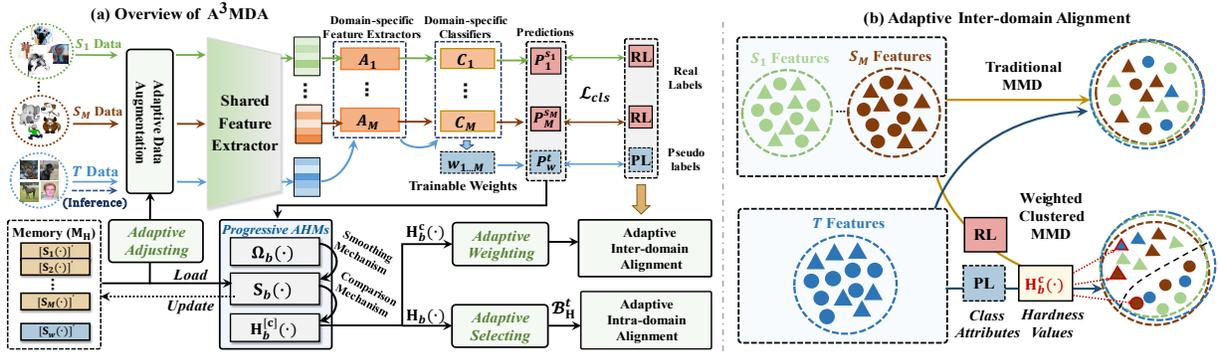

**Fig. 2.** (a) Diagram of our A³MDA framework. Consistent with Fig. 1, **Adaptive Hardness Quantification** derives hardness values from *three* progressive measurements (Basic, Smooth, and Comparative AHM). **Adaptive Hardness Utilization** adaptively leverages them across three scenarios through *three* types of actions (adjusting for data augmentation, weighting for inter-domain alignment, selecting for intra-domain alignment). (b) Illustration of adaptive inter-domain alignment using hardness values from Comparative AHM as weights with class attributes to construct weighted-clustered MMD.

### B. Hardness and Uncertainty-based Measurements

Recently, hardness or uncertainty has gradually emerged as an extra focus for hard sample mining [21], [30] and curriculum learning [31]. For instance, some works [20], [21] leverage the entropy values of target predictions as an uncertainty measurement and minimize them to facilitate domain transfer. Similarly, some other studies [32], [33] endeavor to employ entropy-based hardness analysis on all unlabeled target samples, ranking and categorizing them into hard and easy groups based on a predefined threshold. Furthermore, some recent research has addressed uncertainty regarding label-wise and pair-wise correspondence in other fields, exploring the implications of noisy labels [62]-[64]. For instance, Yang et al. [62] proposed a novel method addressing coupled noisy labels in object Re-ID, effectively rectifying annotation errors. Lin Y. et al. [63] proposed a contrastive matching with momentum distillation for addressing bi-level noisy correspondence in graph matching.

However, most hardness-driven methods in MDA classification tasks primarily rely on direct and instantaneous *entropy measurement*, lacking the involvement of *class attributes (label information)* and *specialized designs (mechanisms)* for diverse scenarios. Such reliance severely restricts the applicability and generalizability of evaluated hardness values. Consequently, despite a few works [34], [35] using model-adaptive hardness strategies in other tasks, most MDA methods only engage sample hardness for qualitative selecting purposes, leaving quantitative hardness applications largely unexplored. As shown in Fig. 1(b), A³MDA addresses the tripartite limitation in MDA with progressive AHMs, with each progress injecting distinct mechanisms to apply hardness values to all given augmentation and alignment scenarios.

### III. METHODOLOGY

MDA aims to transfer knowledge from $M$ labeled source domains $\{S_m\}_{m=1}^M$ to one unlabeled target domain $T$. This corresponds to the $m$-th source batch $\mathcal{B}^{S_m} = \{x_i^{S_m}, y_i^{S_m}\}_{i=1}^{|\mathcal{B}^{S_m}|}$ and target batch $\mathcal{B}^t = \{x_i^t\}_{i=1}^{|\mathcal{B}^t|}$ ($|\mathcal{B}^{S_m}| = |\mathcal{B}^t|$) during training. $x_i^{S_m}$ represents the $i$-th labeled source image and RL = $y_i^{S_m} \in \{0,1\}^K$ denotes its one-hot real label, where $K$ is the number of classes. $x_i^t$ represents the $i$-th unlabeled target image.

As shown in Fig. 2(a), the architecture of A³MDA comprises a shared CNN-based feature extractor $F(\cdot)$, followed by $M$ domain-specific feature extractors $\{A_m(\cdot)\}_{m=1}^M$ and classifiers $\{C_m(\cdot)\}_{m=1}^M$. After retrieving smoothed hardness values from the hardness memory ($M_H$), each $x_i^{S_m}$ and $x_i^t$ will undergo adaptive augmentation and be fed into $F(\cdot)$ to generate features $F_i^{S_m}$ and $F_i^t$. The source feature $F_i^{S_m}$ is then fed into its own ($m$-th) domain-specific modules $A_m(\cdot)$ and $C_m(\cdot)$, producing the aligned feature $F_{m,i}^{S_m}$ and prediction $P_{m,i}^{S_m}$, while the target feature $F_i^t$ is processed by $A_m(\cdot)$ and $C_m(\cdot)$ for every source domain, generating $M$ features $F_{m,i}^t$ and predictions $P_{m,i}^t$. Notably, $M$ predictions are further combined with trainable weights $w_m$ to generate a weighted prediction $P_{w,i}^t = \sum_{m=1}^{M} w_m \cdot P_{m,i}^t$. For a target sample $x_i^t$, $P_{w,i}^t$ is also used to derive its one-hot pseudo-label $PL = \hat{y}_i^t = onehot\left(argmax\left(P_{w,i}^t\right)\right) \cdot 1(z_{\hat{d}} > \tau)$, where $z_{\hat{d}} = \max(P_{w,i}^t)$ is the confidence of the pseudo-class $\hat{d}$, and $\tau$ is the confidence threshold to select pseudo-labels.

### A. Adaptive Hardness Quantification

Prevalent hardness-driven methods simply utilize entropy as a direct measure of hardness, interpreting samples with higher entropy values to have higher levels of difficulty (or uncertainty). Despite proving effectiveness in other unsupervised or supervised classification contexts, most methods solely measure the difficulty of samples at the current epoch and typically employ assessed values as qualitative filters to select hard samples for further processing. Consequently, they struggle to effectively address the diverse scenarios encountered in Multi-source Domain Adaptation (MDA) tasks. To address this, we introduce a series of mechanisms to achieve the *Adaptive Hardness Quantification* for different scenarios. Specifically, we progressively propose *three Adaptive Hardness Measurements* (AHMs) to adaptively quantify the hardness of each labeled source and unlabeled target sample throughout training.

**Basic AHM.** Our progressive AHMs begin with a Basic version, denoted as $\Omega_b(\cdot)$, to gauge the instantaneous hardness in the current epoch. For any source or target sample $x_i^a$ with its prediction $P_{b,i}^a$, $\Omega_b(x_i^a)$ is expressed as:

$$\Omega_b(x_i^a) = \sqrt{\sum_{j=1}^{K} \left(\left[P_{b,i}^a\right]_j\right)^2 \cdot (1 - \delta_{j,z})},$$
$$b = \begin{cases} m, a = s_m \\ w, a = t \end{cases}, z = \begin{cases} k, a = s_m \\ \hat{d}, a = t \end{cases}, \delta_{j,z} = \begin{cases} 1, j = z \\ 0, j \neq z \end{cases}, \quad (1)$$

where $a$ represents the domain in which the sample resides ($s_m$: $m$-th source or $t$: target). $b$ denotes modules that generate predictions ($m$: domain-specific modules of $m$-th source $C_m(A_m(\cdot))$; $w$: passing all modules and weighting their outputs $\sum_{m=1}^{M} w_m \cdot C_m(A_m(\cdot))$). $z$ represents the real class (index) $k$ or the pseudo-class (index) $\hat{d}$. The Kronecker delta $\delta_{j,z}$ is used to ascertain whether index j matches the pseudo or real class index ($z$) of $x_i^a$. In simple terms, the role of Basic AHM is to set the element representing the real class ($k$) in $P_{m,i}^{s_m}$ and the pseudo-class ($\hat{d}$, also the element with the maximum value) in $P_{w,i}^t$ to 0, and then calculates the L2-norm of the remaining elements. To illustrate, for an *m-th* source domain sample $x_i^{s_m}$ with a prediction $P_{m,i}^{s_m} = [0.1\ 0.1\ 0.8]$ produced by domain-specific classifier $C_m(\cdot)$ and a true label $y_i^{s_m} = [0\ 0\ 1]$, the value of $\Omega_m(x_i^{s_m})$ is $\sqrt{(0.1)^2 + (0.1)^2}$, whereas for a more uncertain target sample $x_i^t$ with the weighted prediction $P_{w,i}^t = [0.2\ 0.3\ 0.5]$, we have $\Omega_w(x_i^t) = \sqrt{(0.2)^2 + (0.3)^2} > \Omega_m(x_i^{s_m})$. This implies that more uncertain/harder samples have greater values in $\Omega_m(\cdot)$. Notably, $x_i^t$ utilizes weighted predictions $P_{w,i}^t$ generated by all modules to enable comprehensive handling of target samples.

**Smooth AHM.** Basic AHM may cause instability due to random fluctuations since it only provides assessments of sample hardness in the current epoch. To mitigate this, we introduce Smooth AHM $S_b(\cdot)$ based on $\Omega_b(\cdot)$, which employs Exponential Moving Average (EMA) for temporal smoothing of the evaluated hardness. The formula is:

$$S_b(x_i^a) = \beta \cdot [S_b(x_i^a)]' + (1 - \beta) \cdot \Omega_b(x_i^a), \quad (2)$$

where $[S_b(x_i^a)]'$ represents the smoothed hardness values of $x_i^a$ in the previous epoch (retrieved from memory $M_H$), while $S_b(x_i^a)$ represents the smoothed values of the current epoch. By giving a higher weight (smoothing factor $\beta$) to the previous $[S_b(x_i^a)]'$ and lower weight to the current $\Omega_b(x_i^a)$, $S_b(\cdot)$ focuses more on long-term trends and changes in sample hardness. Notably, $M_H$ will be also updated with the current $S_b(x_i^a)$ to participate in the next training epoch.

**Comparative AHM.** Smooth AHM guarantees the stability of evaluated hardness values. Nonetheless, this absolute stability might not be apt for alignment constraints that only involve batch-wise updates for optimization. This is because using the absolute hardness measurement $S_b(\cdot)$ in constraints can result in certain samples dominating or being marginalized within particular batches, thereby affecting the overall alignment process. To mitigate this concern, we introduce the Comparative AHM, which creates a competitive environment for assessing the relative hardness within each batch or batch-wise cluster. The formula is:

$$H_b^{[c]}(x_i^a) = \frac{S_b(x_i^a)}{\sum_{x_j^a \in \mathcal{B}_{[k/\hat{d}]}^a} S_b(x_j^a)}, \quad (3)$$

where $[c]$ denotes the decision to introduce cluster-wise comparison alongside batch-wise comparison. When represented as the cluster-wise version $H_b^c(\cdot)$, it indicates the selection of samples with the same real class (e.g., $k$) from the source batch $\mathcal{B}^{s_m}$ or the same pseudo-class (e.g., $\hat{d}$) from the target batch $\mathcal{B}^t$. This selection creates the corresponding $\mathcal{B}_k^{s_m}$ and $\mathcal{B}_{\hat{d}}^t$ for dual-level comparisons. While in the default version $H_b(\cdot)$, all samples from $\mathcal{B}^{s_m}$ or $\mathcal{B}^t$ are used.

The progressive design in AHMs embodies our pursuit of Adaptive Hardness Quantification, greatly extending the usability of the measured values. Therefore, our focus is not confined solely to qualitative analysis but also enables us to adaptively utilize hardness across various MDA scenarios.

### B. Adaptive Hardness Utilization

With the acquired progressive hardness values, our attention shifts towards the adaptive utilization of the introduced mechanisms to align with the required properties for diverse MDA scenarios. This necessitates the selection of the most suitable AHM for estimation and adaptive structuring when implementing augmentation and constraints. Fundamentally, these AHMs exhibit an accordant positive correlation, with higher values indicating increased sample hardness and uncertainty. In $A^3$MDA, our guideline of Adaptive Hardness Utilization is that: for harder samples, we should adaptively alleviate the intensity of data Augmentation while reinforcing the potency of domain Alignment. This guideline is realized through *three Actions - Adjusting, Weighting, and Selecting*.

**Adaptive Data Augmentation - Adjusting.** Existing MDA methods typically involve either randomly selecting or using a fixed set of predefined strong augmentations, which are then applied to weakly augmented samples. However, as highlighted in [16], using sample-agnostic strong augmentations can potentially disrupt data distributions in early training stages. Particularly in the MDA context, disregarding the diversity and learning difficulties can inevitably lead to excessive augmentation on already hard-to-train samples. Consequently, some methods [2], [36], despite the increased training burden, resort to augmentations using mutual techniques (e.g., Mixup [37] and CutMix [38]) or searching techniques (e.g., AutoAugment [39] or RandAugment [40]). In contrast, A³MDA continues to rely on individual strong augmentations, but leverages smoothed hardness to adaptively modulate their intensity levels by:

$$h = \begin{cases} [\mathbf{S}_b(x_i^{s_m})]', a = s_m \\ [\mathbf{S}_w(x_i^{t})]', a = t \end{cases}$$

$$\mathcal{A}_{\mathcal{A}}(x_i^{s_m}) = h \cdot \mathcal{A}_{\mathcal{W}}(x_i^{s_m}) + (1-h) \cdot \mathcal{A}_{\mathcal{S}}(x_i^{s_m}), \quad (4)$$

where the temporary variable $h$ denotes the hardness values (i.e. $[\mathbf{S}_b(x_i^{s_m})]'$ and $[\mathbf{S}_w(x_i^t)]'$) previously measured by Smooth AHM in Eq. (2), retrieved from the memory $\mathbf{M}_H$ using the image indexes of $x_i^{s_m}$ and $x_i^t$. $\mathcal{A}_{\mathcal{W}}(\cdot)$ represents weak augmentations, including random scaling, flipping, and cropping, and $\mathcal{A}_{\mathcal{S}}(\cdot)$ encompasses intensity-based strong augmentations, such as invert, blur, contrast, and color jittering, etc., employing settings identical to [16]. We found that this categorization of strong and weak augmentations, beneficial for segmentation, also contributes positively to our application in the MDA classification with adaptive augmentation strategies. The final augmentation, denoted as $\mathcal{A}_{\mathcal{A}}(\cdot)$, adaptively adjusts the intensity level by scaling the proportions of $\mathcal{A}_{\mathcal{S}}(\cdot)$ and its $\mathcal{A}_{\mathcal{W}}(\cdot)$ counterpart. This method shields harder samples from excessive perturbation during early training stages, while also allowing easier, well-fitted samples to benefit from their strongly augmented versions as the model progresses. Notably, the introduced smoothing mechanism in $\mathbf{S}_b(\cdot)$ enables $\mathcal{A}_{\mathcal{A}}(\cdot)$ to progressively raise its intensity levels for various samples, thereby better adapting to the model's evolving generalization ability.

**Adaptive Inter-domain Alignment - Weighting.** A popular strategy in MDA is to minimize inter-domain shifts by imposing discrepancy-based constraints [5], [41], which involves utilizing metrics such as Maximum Mean Discrepancy (MMD) [13], [14] to measure the distance between the distributions of the source and target domains. The formula for traditional MMD loss is:

$$\mathcal{L}_{mmd} = \left\| \phi_i^{s_m} - \phi_i^t \right\|_{\mathcal{H}}^2,$$
$$\overline{\phi^a} = \frac{1}{|\mathcal{B}^a|} \sum_{i=1}^{|\mathcal{B}^a|} \phi_i^a, \quad (5)$$

where $\phi_i^{s_m}$ and $\phi_i^t$ are feature mappings from input samples $x_i^{s_m}$ and $x_i^t$ to a reproducing kernel $\mathcal{H}$ilbert space. $\overline{\phi^{s_m}}$ and $\overline{\phi^t}$ denote their corresponding means.

Despite achieving satisfactory results in many cases, traditional MMD still exhibits limitations in MDA scenarios. To clarify, we categorize it and similar constraints as *'sample-level'*, as they primarily achieve a coarse alignment between source and target distributions by pulling each source and target sample closer. However, such sample-level constraints indiscriminately average distances between each source and target sample, even when they belong to different classes. This imprecise alignment hinders accurate predictions.

To address these limitations, we introduce *'cluster-level'* constraints, aiming to *achieve a finer alignment by adaptively fostering intra-class convergence while promoting inter-class divergence*. As shown in Fig. 2(b), we implement this constraint by combining $\mathcal{L}_{mmd}$ with sample-specific attributes (category and hardness), resulting in a weighted-clustered variant $\mathcal{L}_{mmd}^{WC}$ for adaptive inter-domain alignment. Specifically, class attributes use the real label $y_i^{s_m}$ with index $k$ for the source sample $x_i^{s_m}$ and pseudo-label $\hat{y}_i^t$ with index $\hat{d}$ for target sample $x_i^t$. As for hardness values, unlike data augmentations that use Smooth AHM to gradually adjust the intensity level, we employ cluster-wise Comparative AHM $\mathbf{H}_b^c(\cdot)$ from Eq. (3) to adaptively evaluate the relative importance within batch-wise clusters, assigning greater weights to harder samples. The expression is:

$$\overline{\phi_{\mathbf{H},k}^{s_m}} = \frac{1}{|\mathcal{B}_k^{s_m}|} \sum_{i=1}^{|\mathcal{B}_k^{s_m}|} \phi_{\mathbf{H},i}^{s_m}, \phi_{\mathbf{S},i}^{s_m} = \mathbf{H}_m^c(x_i^{s_m}) \cdot \phi_i^{s_m},$$

$$\overline{\phi_{\mathbf{H},\hat{d}}^t} = \frac{1}{|\mathcal{B}_{\hat{d}}^t|} \sum_{i=1}^{|\mathcal{B}_{\hat{d}}^t|} \phi_{\mathbf{H},i}^t, \phi_{\mathbf{H},i}^t = \mathbf{H}_w^c(x_i^t) \cdot \phi_i^t, \quad (6)$$

$$\mathcal{L}_{mmd}^{WC}(p^{s_m}, p^t) = \sum_{k=1}^{K_c} \left[ \left\| \overline{\phi_{\mathbf{H},k}^{s_m}} - \overline{\phi_{\mathbf{H},\hat{d}=k}^t} \right\|_{\mathcal{H}}^2 - \left\| \overline{\phi_{\mathbf{H},k}^{s_m}} - \overline{\phi_{\mathbf{H},\hat{d}\neq k}^t} \right\|_{\mathcal{H}}^2 \right],$$

where $\phi_i^{s_m}$ and $\phi_i^t$ are further enhanced to $\phi_{\mathbf{S},i}^{s_m}$ and $\phi_{\mathbf{H},i}^t$ by incorporating sample-specific weights generated by $\mathbf{H}_b^c(\cdot)$. The former term in $\mathcal{L}_{mmd}^{WC}$ is to minimize when $x_i^{s_m}$ and $x_i^t$ have the same category attributes $\hat{d} = k$, and the latter term is to maximize when $\hat{d} \neq k$. $K_C$ represents the set of common classes between $\mathcal{B}_k^{s_m}$ and $\mathcal{B}_{\hat{d}}^t$.

With the dual-level comparisons in $\mathbf{H}_b^c(\cdot)$, $\mathcal{L}_{mmd}^{WC}$ can automatically balance the batch-wise importance among different samples with the same class, thus enhancing the batch-based optimization. The final Inter loss combines $\mathcal{L}_{mmd}$ and $\mathcal{L}_{mmd}^{WC}$ for both coarse and fine inter-domain alignments:

$$\mathcal{L}_{Inter}(p^{s_m}, p^t) = (\mathcal{L}_{mmd} + \mathcal{L}_{mmd}^{WC})(p^{s_m}, p^t). \quad (7)$$

**Adaptive Intra-domain Alignment - Selecting.** As highlighted in [32] and [42], comprehensively predicting target labels based on a categorizable target feature space is crucial for enhancing performance. To achieve this, some methods [5], [6], [11], [43] utilize predictions from all domain-specific classifiers to generate a weighted prediction $P_{w,i}^t = \sum_{m=1}^M w_m \cdot P_{m,i}^t$, with trainable weights $w_m$ dictate the reliance on various sources. Despite evident improvements over average predictions, such comprehensiveness remains vulnerable to challenges posed by the intra-domain shift. The intra-domain shift stems from the noise and inconsistencies within the target data distributions, particularly evidenced by certain hard samples that exhibit more resemblance to samples

---

**Algorithm 1 A³MDA algorithm in a mini-batch.**

---

**Input:** Current Epoch ($n$), labeled source batch $\mathcal{B}^{Sm}$, labeled target batch $\mathcal{B}^t$, Hardness Memory $\mathbf{M_H}$.

**Parameters:** Smoothing factor $\beta$, threshold $\tau$, selection ratio $R$, Inter loss weight $\lambda_1$, Intra loss weight $\lambda_2$.

**for** $\{x_i^{Sm}, y_i^{Sm}\} \in \mathcal{B}^{Sm}$ and $\{x_i^t\} \in \mathcal{B}^t$ **do**

    Augment $x_i^{Sm}, x_i^t$ with $M_H$ by Eq. (4).

    Derive mapping for $\phi_i^{Sm}, \phi_i^t$ and predictions $P_{m,i}^{Sm}, P_{m,i}^t$.

    Derive weighted prediction $P_{w,i}^t$ and pseudo-label $\hat{y}_i^t$.

    Calculate Basic AHM $\mathbf{\Omega}_m(x_i^{Sm}), \mathbf{\Omega}_w(x_i^t)$ by Eq. (1).

    Calculate Smooth AHM: $\mathbf{S}_m(x_i^{Sm}), \mathbf{S}_w(x_i^t)$ and update $\mathbf{M_H}$ with $\beta$ by Eq. (2).

**end for**

Calculate Comparative AHM: $\mathbf{H}_m^c(x_i^{Sm}), \mathbf{H}_w^c(x_i^t), \mathbf{H}_w(x_i^t)$ by Eq. (3).

Create $\mathcal{B}_k^{Sm}, \mathcal{B}_{a=k}^t, \mathcal{B}_{d\neq k}^t$ for each class $k$. Calculate $\mathcal{L}_{Inter}$ by Eqs. (5) to (7).

Create $\mathcal{B}_\mathbf{H}^t$ by selecting top ($R\%$) $\mathbf{H}_w(t)$ values in $\mathcal{B}^t$. Calculate $\mathcal{L}_{Intra}$ by Eqs. (8) to (9).

Calculate $\mathcal{L}_{cls}$ using $< P_{m,i}^{Sm}, y_i^{Sm} >$ and $< P_{w,i}^t, \hat{y}_i^t >$.

**return** $\lambda_1 \cdot \mathcal{L}_{Inter} + \lambda_2 \cdot \mathcal{L}_{Intra} + \mathcal{L}_{cls}$ by Eq. (10).

---

from other classes than their own. These hard samples can cause over-reliance on certain source domains in the weighted prediction, potentially leading to erroneous pseudo-labels and the distortion of the target feature space.

To reduce the intra-domain shift, A³MDA introduces another *cluster-level* constraint based on weighted predictions of hard samples. Similar to $\mathcal{L}_{mmd}^{WC}$, the constraint for intra-domain alignment also integrates a dual-level comparison mechanism, but is implemented in two operations: (1) an external batch-wise selection of harder samples using the rankings of default Comparative AHM $\mathbf{H}_b(\cdot)$, and (2) an internal cluster-wise competition using a pseudo-contrastive matrix. The rationale for not directly using the cluster-wise version $\mathbf{H}_b^c(\cdot)$ is that employing it as weights in a purely unsupervised constraint is unfeasible due to its dependence on relatively reliable pseudo-labels, and utilizing it to select harder samples may also yield an unstable prioritization, especially in batch-wise clusters with limited samples.

As shown in Fig. 3, the batch-wise selection of harder samples is achieved by sorting all target samples in batch $\mathcal{B}^t$ based on their $\mathbf{H}_b(\cdot)$ (also $\mathbf{H}_w(\cdot)$) rankings, and retaining the top $R$ (selection ratio) to form a new batch, denoted as $\mathcal{B}_\mathbf{H}^t$. Then, we create a Pseudo-Label Matrix (PLM) $\text{PL}_\mathcal{B}^2 = \text{PL}_\mathcal{B}$. $\text{PL}_\mathcal{B}^T$ to establish batch-wise positive/negative relationships, where $\text{PL}_\mathcal{B} \in R^{|\mathcal{B}_\mathbf{H}^t| \times K}$ combines the pseudo-labels of all target samples in $\mathcal{B}_\mathbf{H}^t$, and $\text{PL}_\mathcal{B}^T$ represents its transposition. Here, $\left(\text{PL}_\mathcal{B}^2\right)_{i,j} = 1/0$ indicates whether the $i$-th and $j$-th samples in the target batch $\mathcal{B}^t$ are positive (1: with the same pseudo-class) or negative (0: with different pseudo-class) pairs. For diagonal elements in $\text{PL}_\mathcal{B}^2$, their values are always 1, as each sample is positive with itself; for non-diagonal elements, their values depend on whether two samples share the same pseudo-class.

After establishing PLM, our attention turns to compute pairwise similarities by constructing diverse views of samples.

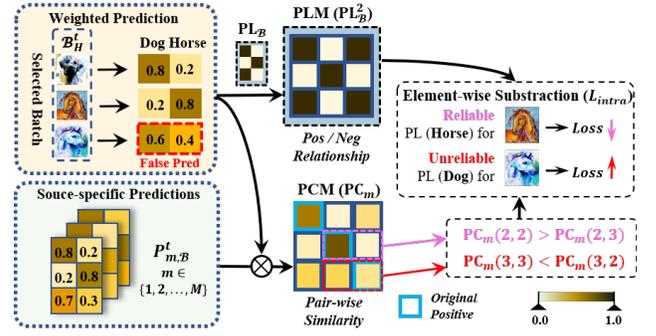

**Fig. 3.** Illustration of adaptive intra-domain alignment. Target batch $\mathcal{B}_\mathbf{H}^t$ is selected based on the sorting of Comparative AHM values. The intra-domain loss is then built by element-wise subtraction of PLM and PCM.

This is achieved via a Pseudo Contrastive Matrix (PCM). In contrast to the traditional contrastive learning, which augments anchors to create a single positive view and projects these views into feature vectors, PCM directly selects weighted predictions as anchors and uses predictions with the same or different pseudo-classes, generated by domain-specific classifiers, as positive or negative views. The PCM for the $m$-th source is represented as:

$$(PC_m)_{i,j} = \frac{exp\left(P_{w,i}^t \cdot P_{m,j}^t / tem\right)}{\sum_{x_k^t \in \mathcal{B}_\mathbf{H}^t} exp\left(P_{w,i}^t \cdot P_{m,k}^t / tem\right)}, \quad (8)$$

where $(PC_m)_{i,j}$ indicates the pair-wise similarity between the anchor ($P_{w,i}^t$: weighted view of the $i$-th sample) and its positive/negative views ($P_{m,j}^t$: domain-specific views of the $j$-th sample). Here, the domain-specific view $P_{m,i}^t$ are named as *original positives* of the anchor as they originate from the same $i$-th sample. Notably, in addition to the dot product for similarity measurement, we adopt the softmax function with the temperature parameter $tem = 0.15$ to obtain the final values of each row. This allows views of other samples to compete with *original positives* to become "more positive" to the anchor. By minimizing the L1 distance between PLM and PCM, the Intra loss is formulated as:

$$\mathcal{L}_{Intra}(p^{Sm}, p^t) = \text{Mean}\left(\left| PL_\mathcal{B}^2 - PC_m \right|\right), \quad (9)$$

where $\text{Mean}(\cdot)$ computes the mean value across all elements in the matrix. The intrinsic mechanisms of $\mathcal{L}_{Intra}$ are described as follows: (1) If samples i and j belong to different pseudo-classes, Eq. (9) for the i-th sample can be expressed as the term $\mathcal{L}_{intra(i,j)} = (1 - (PC_m)_{i,i}) + (PC_m)_{i,j}$. In this case, if $(PC_m)_{i,i} < (PC_m)_{i,j}$, it suggests that the original positive ($P_{m,i}^t$) lose in competition with the view of the j-th sample ($P_{m,j}^t$), implying an uncertain pseudo-label of i-th sample. Consequently, this term results in a larger loss value to correct this potentially erroneous pseudo-label, with the pseudo-class represented by the j-th sample potentially becoming the true class. Similarly, if $(PC_m)_{i,i} > (PC_m)_{i,j}$, this indicates that original positives maintain their dominant position, suggesting a relatively reliable pseudo-label for the i-th sample. (2) Conversely, if i and j belong to the same pseudo-class, Eq. (9) takes the form $\mathcal{L}_{intra(i,j)} = (1 - (PC_m)_{i,i}) + (1 - (PC_m)_{i,j})$.

TABLE I
COMPARISON OF CLASSIFICATION ACCURACY (%) ON OFFICE-31 AND OFFICE-HOME.

| Datasets | | Office-31 | | | | Office-Home | | | |
|---|---|---|---|---|---|---|---|---|---|
| Protocols | Methods | →D | →W | →A | Avg | →A | →C | →P | →R | Avg |
| Single Best | Source-only | 99.3 | 96.7 | 62.5 | 86.2 | 65.3 | 49.6 | 79.7 | 75.4 | 67.5 |
| | DAN [13] | 99.5 | 96.8 | 66.7 | 87.7 | 68.2 | 56.5 | 80.3 | 75.9 | 70.2 |
| | DANN [2] | 99.1 | 96.9 | 68.2 | 88.1 | 67.9 | 55.8 | 80.3 | 76.3 | 69.3 |
| Source Combine | DAN [13] | 99.6 | 97.8 | 67.6 | 88.3 | 68.5 | 59.4 | 79.0 | 82.5 | 72.4 |
| | DANN [2] | 99.7 | 98.1 | 67.6 | 88.5 | 68.4 | 59.1 | 79.5 | 82.7 | 72.4 |
| | D-CORAL [1] | 99.3 | 98.0 | 67.1 | 88.1 | 68.1 | 58.6 | 79.5 | 82.7 | 72.2 |
| Multi-Source | M³SDA [5] | 99.3 | 98.0 | 67.2 | 88.2 | 66.2 | 58.6 | 79.5 | 81.4 | 71.4 |
| | MFSAN [18] | 99.5 | 95.5 | 72.7 | 90.2 | 72.1 | 62.0 | 80.3 | 81.8 | 74.1 |
| | MIAN [4] | 99.5 | 98.5 | 74.7 | 90.9 | 69.4 | 63.1 | 79.6 | 80.4 | 73.1 |
| | T-SVDNet [44] | 99.4 | 99.6 | 74.1 | 91.0 | 71.9 | 65.1 | 82.6 | 81.8 | 75.3 |
| | SPS [45] | **100.0** | 99.3 | 73.8 | 91.0 | <u>75.1</u> | <u>66.0</u> | 84.4 | 84.2 | <u>77.4</u> |
| | DFSE [46] | 99.4 | 98.8 | 73.2 | 90.5 | 73.4 | 62.7 | <u>84.5</u> | <u>85.3</u> | 76.5 |
| | MIEM [50] | <u>99.8</u> | <u>99.7</u> | <u>75.9</u> | <u>91.7</u> | 73.6 | 65.9 | 83.2 | 83.1 | 76.5 |
| | A³MDA (ours) | **100.0** | **99.8** | **77.9** | **92.6** | **75.4** | **66.3** | **85.6** | **85.4** | **78.2** |

In this case, both $(PC_m)_{i,i}$ and $(PC_m)_{i,j}$ are optimized to have larger values to minimize this loss term.

In summary, when the PLM ($PL_B^2$) exhibits relatively smaller distances with PCM from different source-specific views ($PC_m$), the selected pseudo-labels for hard samples are more reliable. This process effectively rectifies the false pseudo-labels of those hard samples and largely avoids the excessive reliance on certain domain-specific predictions (views) in weighted predictions.

**Training and Inference.** During training, we employ cross-entropy loss $\mathcal{L}_{cls}$ for all data predictions and their real/pseudo labels. By combining $\mathcal{L}_{Inter}$ and $\mathcal{L}_{Intra}$ using hyperparameters $\lambda_1$ and $\lambda_2$ for $M$ source domains, the final objective function is:

$$\mathcal{L}_{total} = \sum_{m=1}^{M}(\lambda_1 \cdot \mathcal{L}_{Inter} + \lambda_2 \cdot \mathcal{L}_{Intra} + \mathcal{L}_{cls}), \quad (10)$$

For better understanding, we summarize the batch-wise training process in Algorithm 1. As for inference in the target domain, we directly utilize the weighted predictions as the final prediction results.

## IV. EXPERIMENTS

To validate the generalizability and superiority of our A³MDA, we conduct comprehensive evaluations in this section using seven widely used publicly available datasets and compare the results with state-of-the-art methods.

### A. Datasets

**Office-31** [47] is a commonly used MDA dataset, with 4,110 images in 31 categories from three domains: Amazon (A), Webcam (W), and DSLR (D). This dataset is imbalanced, with 2,817, 795, and 498 images in domains A, W, and D, respectively. **Office-Home** [48] is a benchmark dataset for MDA tasks, containing 15,588 images across Artistic (A), Clip Art (C), Product (P), and Real-World (R) domains, covering 65 classes in total. **DomainNet** [5] is a challenging dataset with 345 categories and 6 domains, including Clipart (Clp), Infograph (Inf), Painting (Pnt), Quickdraw (Qdr), Real (Rel), and Sketch (Skt). **ImageCLEF-DA** [14] is a dataset with 12 categories shared by 3 public domains: Caltech-256 (C), ImageNet ILSVRC 2012 (I), and Pascal VOC 2012 (P). Each domain contains 600 images, and each category has 50 images. **PACS** [51] is a dataset that includes images from four domains: Art (A), Cartoon (C), Sketch (S), and Photo (P). Each domain contains 7 categories, with the following number of images: P (1,670 images), A (2,048 images), C (2,048 images), and S (3,929 images). **Digits-5** [5] dataset consists of handwritten digit images from five different domains: MNIST-M (mm), MNIST (mt), USPS (up), SVHN (sv), and SYN (syn). There are 10 classes in total, corresponding to the digits 0 to 9. **Office-Caltech** [52] dataset comprises four different domains: Webcam (W), DSLR (D), Caltech10 (C), and Amazon (A). The number of images in each domain is as follows: Webcam (157 images), DSLR (295 images), Caltech10 (1,123 images), and Amazon (958 images) with each domain containing 10 categories.

TABLE II
COMPARISON OF CLASSIFICATION ACCURACY (%) ON DOMAINNET.

| Methods | →Clp | →Inf | →Pnt | →Qdr | →Rel | →Skt | Avg |
|---|---|---|---|---|---|---|---|
| M³SDA [5] | 58.6 | 26.0 | 52.3 | 6.3 | 62.7 | 49.5 | 42.6 |
| T-SVDNet [44] | 66.1 | 25.0 | 54.3 | 16.5 | 65.4 | 54.6 | 47.0 |
| PTMDA [41] | 66.0 | 28.5 | 58.4 | 13.0 | 63.0 | 54.1 | 47.2 |
| DSFE [46] | 68.2 | 25.8 | 58.8 | 18.3 | 71.9 | 57.6 | 50.1 |
| SPS [45] | 70.8 | 24.6 | 55.2 | 19.4 | 67.5 | 57.6 | 49.2 |
| MIEM [50] | 69.0 | 28.6 | 58.7 | <u>20.5</u> | 68.9 | 59.2 | 50.8 |
| PMSDAN [72] | 69.4 | <u>28.4</u> | 59.2 | 18.3 | <u>72.2</u> | 58.6 | 51.0 |
| MCC-DA [73] | **74.3** | 26.1 | <u>60.2</u> | 19.4 | 71.7 | **60.8** | <u>52.1</u> |
| A³MDA(ours) | <u>71.4</u> | **31.6** | **60.3** | **21.1** | **73.3** | <u>60.7</u> | **53.1** |

### B. Implementation and Training Details

When implementing A³MDA, we employ diverse backbone networks (denoted as $F(\cdot)$) for different datasets to ensure consistency with other comparative methods. Specifically, for the Office-Caltech and DomainNet datasets, we utilize ResNet101 [53]. For the Digits-5 dataset, we adhere to the architecture proposed in M³SDA [5]. For the remaining datasets, we use ResNet50. All these backbone models are pretrained on ImageNet. For the domain-specific feature extractor $A_m(\cdot)$, we apply the (Conv1×1, Conv3×3, and Conv1×1) structure to reduce the number of channels from 2048 to 256. As for the classifier $C_m(\cdot)$, it comprises a single fully connected layer that maps features with 256 channels from $A_m(\cdot)$ to the class space of the respective dataset.

TABLE III
COMPARISON OF CLASSIFICATION ACCURACY (%) ON IMAGECLEF-DA.

| Protocols | Methods | →P | →C | →I | Avg |
|-----------|---------|-----|-----|-----|-----|
| Single Best | Source-only | 74.8 | 91.5 | 83.9 | 83.4 |
| | DAN [13] | 75.0 | 93.3 | 86.2 | 84.8 |
| | D-CORAL [1] | 76.9 | 93.6 | 88.5 | 86.3 |
| Source Combine | DAN [13] | 77.6 | 93.3 | 92.2 | 87.7 |
| | DANN [2] | 77.9 | 93.7 | 91.8 | 87.8 |
| Multi-Source | M³SDA [5] | 77.3 | 94.3 | 91.9 | 88.0 |
| | MFSAN [18] | 79.1 | 95.4 | 93.6 | 89.4 |
| | MIAN [4] | 77.6 | 95.1 | 91.5 | 88.1 |
| | T-SVDNet [44] | 78.9 | 95.5 | 93.7 | 89.3 |
| | PTMDA [41] | <u>79.1</u> | <u>97.3</u> | 94.1 | <u>90.2</u> |
| | DSFE [46] | 78.7 | 96.0 | 93.5 | 89.4 |
| | MIEM [50] | 79.0 | 97.1 | <u>94.3</u> | 90.1 |
| | A³MDA(ours) | **79.7** | **97.8** | **97.5** | **91.7** |

TABLE IV
COMPARISON OF CLASSIFICATION ACCURACY (%) ON PACS.

| Protocols | Methods | →A | →C | →S | →P | Avg |
|-----------|---------|-----|-----|-----|-----|-----|
| Source Combine | Source-only | 85.16 | 76.78 | 71.22 | 97.94 | 82.77 |
| | DAN [13] | 87.35 | 83.92 | 77.07 | 98.32 | 86.67 |
| | DANN [2] | 87.53 | 84.21 | 78.44 | 97.64 | 86.96 |
| Multi-Source | M³SDA [5] | 84.20 | 85.68 | 74.62 | 94.47 | 84.74 |
| | MFSAN [18] | 90.19 | 90.47 | 81.53 | 97.23 | 89.85 |
| | MIAN [4] | 90.32 | 88.42 | 81.23 | <u>98.71</u> | 89.67 |
| | T-SVDNet [44] | 91.39 | <u>91.39</u> | 84.97 | 97.93 | 91.42 |
| | DSFE [46] | 90.43 | 90.71 | 85.59 | 98.54 | 91.33 |
| | MIEM [50] | <u>91.87</u> | 90.44 | <u>88.07</u> | <u>98.71</u> | <u>92.27</u> |
| | A³MDA(ours) | **92.14** | **91.56** | **90.35** | **99.30** | **93.34** |

TABLE V
COMPARISON OF CLASSIFICATION ACCURACY (%) ON DIGITS-5.

| Methods | →mm | →mt | →up | →sv | →syn | Avg |
|---------|-----|-----|-----|-----|------|-----|
| M³SDA [5] | 72.8 | 98.4 | 96.1 | 81.3 | 89.6 | 87.7 |
| MDDA [6] | 78.6 | 98.8 | 93.9 | 79.3 | 79.3 | 88.1 |
| LtC-MSDA [42] | 85.6 | 99.0 | 98.3 | 83.2 | 93.0 | 91.8 |
| STEM [29] | 89.7 | **99.4** | 98.4 | 89.9 | **97.5** | 95.0 |
| DIDA-Net [9] | 85.7 | 99.3 | 98.6 | **91.7** | 97.3 | 94.5 |
| A³MDA(ours) | **92.2** | 99.3 | **98.7** | 91.4 | 96.2 | **95.6** |

TABLE VI
COMPARISON OF CLASSIFICATION ACCURACY (%) ON OFFICE-CALTECH.

| Protocols | Methods | →A | →C | →D | →W | Avg |
|-----------|---------|-----|-----|-----|-----|-----|
| Source Combine | Source-only | 86.1 | 87.8 | 98.3 | 99.0 | 92.8 |
| | DAN [13] | 94.8 | 89.7 | 98.2 | 99.3 | 95.5 |
| | DANN [2] | 94.8 | 89.7 | 98.2 | 99.3 | 95.5 |
| Multi-Source | M³SDA [5] | 94.5 | 92.2 | 99.2 | 98.9 | 96.4 |
| | MFSAN [18] | 95.4 | 93.8 | <u>99.4</u> | <u>99.7</u> | 97.1 |
| | MIAN [4] | 96.1 | 94.6 | 99.0 | 99.3 | 97.2 |
| | T-SVDNet [44] | 96.6 | 93.9 | **100.0** | 99.5 | 97.5 |
| | STEM [29] | **98.4** | 94.2 | **100.0** | **100.0** | <u>98.2</u> |
| | DSFE [46] | 95.4 | 94.6 | <u>99.4</u> | 99.5 | 97.2 |
| | MIEM [50] | 96.4 | <u>96.0</u> | **100.0** | 99.5 | 98.0 |
| | A³MDA(ours) | <u>97.3</u> | **97.7** | **100.0** | **100.0** | **98.7** |

When training A³MDA, we set the learning rate to 0.001 for the backbone $F(\cdot)$ pretrained on ImageNet, and 0.01 for the domain-specific modules $A_m(\cdot)$ and $C_m(\cdot)$ trained from scratch. Training lasts 200 epochs on an RTX 3090 GPU, following the optimizer and learning schedule in [18]. We maintain a fixed random seed of 10 over 3 runs and report the average results. We adjust $\lambda_1$ for $\mathcal{L}_{Inter}$ asymptotically from 0 to 1 by the formula $\frac{2}{exp(-\theta p)} - 1$ in DANN [2], where $\theta = 10$ and $p$ is linearly changing from 0 to 1 as training iteration increases. Based on our trial study, we set $\lambda_2$ for $\mathcal{L}_{Intra}$ as 0.7 and the pseudo-label threshold $\tau$ as 0.6. For the smoothing factor $\beta$ in Smooth AHM, and the selection ratio $R$ to filter harder samples, we set them to 0.8 and 0.4. The batch size ($B$) for all datasets is set to 32.

## C. Comparative Experiments

In this section, we compare A³MDA with state-of-the-art single-source domain adaptation (SDA) and multi-source domain adaptation (MDA) algorithms. Specifically, two protocols are adopted to train SDA methods, including 1)

Single Best, which reports the best result among all source domains, and 2) Source Combination, which naively combines all source domains and then performs single-source domain adaptation, while one protocol 3) Multi-Source for MDA methods. Source-Only refers to directly transferring the model trained in source domains to the target domain. Here, we select DAN [13], D-CORAL [1], and DANN [2], for SDA methods, and M³SDA [5], MFSAN [18], MDDA [6], MIAN [4], T-SVDNet [44], PTMDA [41], DSFE [46], SPS [45], MIEM [50], LtC-MSDA [42], STEM [29], DIDA-Net [9], PMSDAN [72], and MCC-DA [73] for MDA methods. To ensure fairness, we either quote the results of compared methods from their papers or reproduce them using the released codes if the results on a specific dataset are not available. For all compared methods on each dataset, we maintain consistency by employing the same backbone architecture ($F(\cdot)$ in A³MDA) and data pre-processing routines. Notably, as some methods may excel on one dataset but not on others, we may introduce slight variations in the selection of compared methods for each dataset. To better highlight the superiority of our method, we use **bold** for the best results and <u>underline</u> for the second-best results when comparing accuracy across different datasets in Tables I to VI.

**Classification Accuracy.** We first compared the accuracy on each task across seven datasets. The results on **Office-31** and **Office-Home** are shown in Table I. For Office-31, our method attains the highest average accuracy of 92.6%, surpassing the second-best competitor MIEM by 0.9%. As for the Office-Home dataset, our proposed A³MDA also outperforms all comparative methods, surpassing the second-best method SPS by 0.8%.

TABLE VII
COMPARISON OF COMPUTATIONAL COMPLEXITY.

| Datasets (Scale, Arc) | DomainNet (Large, R101) | | | Office-Home (Middle, R50) | | |
|---|---|---|---|---|---|---|
| Methods | MACs | Time | Acc | MACs | Time | Acc |
| $M^3SDA$ [5] | 10.45 | 16.31 | 42.6 | 6.76 | 6.56 | 71.4 |
| T-SVDNet [44] | 10.21 | 16.56 | 47.0 | 6.69 | 6.60 | 75.3 |
| DSFE [46] | 10.04 | 15.82 | 50.1 | 6.60 | 6.41 | 76.5 |
| $A^3MDA$ (ours) | 10.69 | 17.73 | **53.1** | 6.96 | 7.08 | **78.2** |

TABLE VIII
ABLATION STUDY ON KEY COMPONENTS FOR AVERAGE
ACCURACY (%) ON OFFICE-31 AND OFFICE-HOME.

| Levels/ Scenarios | Methods | Office-31 | Office-Home |
|---|---|---|---|
| | $\mathcal{L}_{mmd}$ (Baseline) | 88.1 | 73.3 |
| Data Augmentation | + Aug w/o Adjusting | 88.5 | 73.6 |
| | + Aug w/ Adjusting | 89.5 | 74.5 |
| Inter-domain Alignment | + $\mathcal{L}_{mmd}^{WC}$ w/o Weighting | 90.5 | 75.6 |
| | + $\mathcal{L}_{mmd}^{WC}$ w/ Weighting | 91.2 | 76.4 |
| Intra-Domain Alignment | + PCM w/o Selecting | 91.9 | 77.6 |
| | + PCM w/ Selecting | **92.6** | **78.2** |

The results on the most challenging **DomainNet** are shown in Table II. $A^3MDA$ outperforms MCC-DA by an average accuracy of 1.0% due to its weighted clustered MMD loss, which adaptively measures hardness values to achieve precise inter-domain alignment. $A^3MDA$ also surpasses PMSDAN by 2.0% on average, as PMSDAN uses a single-purpose weighting discrepancy, while $A^3MDA$ employs a more sophisticated discrepancy formulation for diverse scenarios. Furthermore, while SPS, MIEM, and MCC-DA focus on pseudo-label strategies, $A^3MDA$ explicitly addresses intra-domain shifts to effectively mitigate erroneous pseudo-labels, resulting in an average performance improvement of 2.4%.

The results on **ImageCLEF-DA** are shown in Table III. $A^3MDA$ excels across all domains, securing the top rank with an average accuracy of 91.7%. This outperforms the second-best method PTMDA by 1.5%. Notably, our method exhibits a substantial 3.2% lead over MIEM on the '→I' task.

Table IV presents the results obtained on the **PACS** dataset. Our method outperforms the second-ranked method MIEM by a margin of 1.07%, achieving superior performance on three out of four tasks with an average accuracy of 93.34% across the four domains.

Table V showcases the results on the **Digits-5** dataset. Our method achieves the highest average accuracy of 95.6%, surpassing the second-best method, STEM, by 0.6%. Particularly, in the '→mm' task, our method demonstrates a significant improvement of 2.5% compared to STEM.

The results on **Office-Caltech** are shown in Table VI. Our method achieves the best performance on three out of four tasks, obtaining an average accuracy of 98.7% across four domains and ranking first in the list.

**Computational Complexity.** We conducted an extensive analysis of the computational cost (measured in MACs and training time in hours) and classification performance (measured in accuracy) of the $A^3MDA$ framework. This comparison was conducted on datasets of varying scales, utilizing different feature extraction backbones: DomainNet (large-scale) with ResNet101 ('R101') and Office-Home (middle-scale) with ResNet50 ('R50'). This was done to evaluate the impact of dataset scale and feature extraction architecture on the computational complexity of each method. Notably, we placed particular emphasis on $M^3SDA$, as it represents the simplest discrepancy-based method, relying solely on general MMD loss for inter-domain alignment.

As observed from Table VII, $A^3MDA$ incurs a modest increase in MACs and training time due to its adaptive hardness strategy, which calculates and applies different hardness measurements across different scenarios. However, this additional cost is deemed acceptable, as the extra computations related to hardness values primarily involve batch-wise predictions rather than high-dimensional features. Moreover, the framework achieves substantial accuracy improvements. Compared to the simplest $M^3SDA$, the increase in MACs and training time for $A^3MDA$ remains under 10%, while the average accuracy improves substantially by 10.5% and 6.8% on the two datasets, showcasing superior performance with minimal complexity escalation.

Overall, across all the datasets mentioned, which cover almost all widely used datasets for MDA classification tasks, $A^3MDA$ consistently demonstrates superior performance on average. Besides, experiment results on computation metrics also indicate that our method can achieve a better trade-off between performance and complexity. This highlights the immense potential of utilizing the proposed adaptive hardness strategy to enhance traditional discrepancy-based methods with minimal additional computational cost.

### E. Ablation Studies

In this section, we conduct a thorough exploration of $A^3MDA$ from various perspectives. This exploration includes the examination of key components and their utilization, the assessment of different hardness measurements in various utilization scenarios, the exploration of hyperparameter settings, and various visualization experiments such as t-SNE, similarity matrix, and trends of hardness values during training.

**Evaluation on Adaptive Hardness Utilization.** To evaluate the effectiveness of various key components and their utilization in $A^3MDA$, we conduct ablative experiments on the Office-31 and Office-Home datasets. We establish traditional MMD as our baseline ($\mathcal{L}_{mmd}$) and gradually incorporate three levels of components. 'Aug' denotes data augmentation, and 'Adjusting' involves using hardness values from Smooth AHM $\mathbf{S}(\cdot)$ for adaptive intensity adjustment. '$\mathcal{L}_{mmd}^{WC}$' represents the weighted clustered MMD for inter-domain alignment, while 'Weighting' denotes the use of Comparative AHM $\mathbf{H^c}(\cdot)$ for sample-specific weights. PCM signifies the inclusion of a pseudo-contrastive matrix for intra-domain alignment, and 'Selecting' indicates whether Comparative AHM $\mathbf{H}(\cdot)$ is employed to choose more challenging samples for PCM formulation. As shown in Table VIII, relying solely on random strong data augmentation does not significantly improve the model due to the presence of over-augmentation.

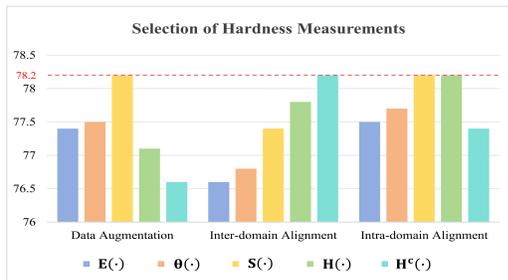

**Fig. 4.** Investigation of native entropy against various AHMs.

However, introducing the adjusting strategy enables dynamic adaptation of augmentation, leading to notable improvements of 1.4% and 2.2% over the baseline on Office-31 and Office-Home, respectively. For inter-domain alignment, we progressively incorporate sample-specific class attributes and hardness values to refine $\mathcal{L}_{mmd}$, leading to the formulation of $\mathcal{L}_{mmd}^{WC}$. This formulation significantly enhances the robustness and precision of the alignment process, contributing to a noteworthy improvement of 1.7% on Office-31 and 1.8% on Office-Home. For intra-domain alignment, integrating PCM improves the categorizability of the target domain feature space by fostering an internal competitive environment. Based on hardness rankings, further injecting a simple yet adaptive selection strategy externally intensifies the competition within PCM, aiding in the correction of erroneous pseudo-labels. This combined enhancement results in a significant 1.4% and 1.8% improvement on Office-31 and Office-Home.

**Evaluation on Adaptive Hardness Quantification.** The core of A³MDA also lies in Adaptive Hardness Quantification across various scenarios. In this study, we benchmark the native Shannon entropy $\mathbf{E}(\cdot) = \sum_k^K p(y = k|x) log(p(y = k|x))$ [72] (treated as another basic AHM) against our Adaptive Hardness Measurements (AHMs): the Basic AHM $\boldsymbol{\theta}(\cdot)$, the Smooth AHM $\mathbf{S}(\cdot)$ with a smoothing mechanism, and two Comparative AHMs ($\mathbf{H}(\cdot)$ and $\mathbf{H^c}(\cdot)$) with single- and dual-level comparisons. To systematically evaluate these AHMs, we conduct a step-by-step comparison across three levels using the Office-Home dataset. For data augmentation, we construct dedicated memory banks for all AHMs to store values from the previous epoch, which are then used in Eq. (4). For inter-domain alignment, we assign different AHMs to source and target samples. For intra-domain alignment, we employ the sorting of different AHMs to construct the pseudo-contrastive matrix (PCM). Importantly, to ensure consistency, we only modify the AHM used at the investigated level, while maintaining the optimal configurations for the other two levels: $\mathbf{S}(\cdot)$ in Data Augmentation, $\mathbf{H^c}(\cdot)$ in inter-domain Alignment, and $\mathbf{H}(\cdot)$ in intra-domain Alignment.

As observed in Fig. 4, for **Data Augmentation**, both the native entropy $\mathbf{E}(\cdot)$ and the basic AHM $\boldsymbol{\theta}(\cdot)$, which lack a smoothing mechanism, perform suboptimal compared to $\mathbf{S}(\cdot)$. This is primarily due to their sensitivity to random fluctuations, as they evaluate sample hardness only within the current epoch. Furthermore, incorporating comparative mechanisms in $\mathbf{H}(\cdot)$ and $\mathbf{H^c}(\cdot)$ over $\mathbf{S}(\cdot)$ also fails to yield satisfactory results when applied to data augmentations, as their effectiveness is heavily contingent on the relative difficulty of other samples within the batch from the previous epoch, making them unsuitable for adaptive augmentation in the current epoch. For **Inter-domain Alignment**, we found that AHMs with comparative mechanisms can automatically balance the batch-wise importance among different samples in the weighted MMD loss, thereby enhancing batch-based optimization. Meanwhile, $\mathbf{H^c}(\cdot)$, which builds upon $\mathbf{H}(\cdot)$ but further introduces cluster-wise comparison, achieves better results by adaptively fostering intra-class convergence while promoting inter-class divergence. For **Intra-domain alignment**, we found that both the native entropy $\mathbf{E}(\cdot)$ and the basic AHM $\boldsymbol{\theta}(\cdot)$ do not perform well due to their inability to stably and reliably perceive the hardness of target samples. Consequently, constructing an effective PCM became challenging. Interestingly, employing $\mathbf{H}(\cdot)$ and maintaining $\mathbf{S}(\cdot)$ both yield optimal performance, as the hardness ranking of samples within batches remained unchanged after batch-wise comparison. However, the further introduction of cluster-wise comparison in $\mathbf{H^c}(\cdot)$ results in the poorest performance. This is because utilizing $\mathbf{H^c}(\cdot)$ to select harder samples may lead to an unstable prioritization of batch-wise clusters with a limited number of target samples, thereby rendering the selection process almost ineffective.

**Exploring Different Backbone Architectures.** We experimented with various backbone architectures on the Office-Home dataset, including InceptionV3 [65], DenseNet161 [66], ViT-B/16 [67], ResNet34 [53], and ResNet101 [53], which vary in complexity and design philosophies. The original A³MDA utilized ResNet50 [53], consistent with prior MDA studies [18, 45, 46]. As detailed in TABLE IX, our method can adapt well to different backbones and achieve competitive performance across all of them. These findings underscore the wide applicability of our method across various extraction backbones.

TABLE X
ABLATION STUDY OF DIFFERENT METRIC DISTANCES FOR AVERAGE ACCURACY (%) ON OFFICE-HOME.

| Methods | →A | →C | →P | →R | Avg |
|---|---|---|---|---|---|
| L2 [68] | 58.7 | 45.8 | 67.6 | 68.8 | 60.2 |
| L2 + Ours | 62.8 | 53.7 | 4 | 74.4 | 66.1 (↑5.9) |
| WAS [69] | 71.1 | 62.0 | 81.1 | 80.3 | 73.6 |
| WAS + Ours | 75.5 | 64.4 | 84.9 | 85.5 | 77.6 (↑4.0) |
| CORAL [1] | 71.4 | 61.9 | 79.7 | 80.7 | 73.4 |
| CORAL + Ours | 75.0 | 64.5 | 85.5 | 85.3 | 78.0 (↑4.6) |
| MMD [18] | 71.1 | 61.9 | 79.3 | 80.8 | 73.3 |
| MMD + Ours | 75.4 | 66.3 | 85.6 | 85.4 | 78.2 (↑4.9) |

TABLE IX
COMPARISON OF DIFFERENT BACKBONES ON OFFICE-HOME.

| Methods | →A | →C | →P | →R | Avg | Params | MACs |
|---|---|---|---|---|---|---|---|
| InceptionV3 | 77.1 | 66.7 | 85.3 | 85.6 | 78.7 | 29.87M | 8.42G |
| DensNet161 | 79.3 | 67.5 | 85.7 | 86.1 | 79.7 | 31.42M | 10.75G |
| ViT-B/16 | 79.0 | 67.9 | 86.1 | 86.8 | 80.0 | 90.31M | 20.68G |
| ResNet34 | 74.2 | 65.0 | 83.2 | 83.1 | 76.4 | 24.14M | 5.70G |
| ResNet101 | 77.8 | 67.8 | 85.5 | 85.5 | 79.2 | 47.29M | 10.69G |
| A³MDA (ours) | 75.4 | 66.3 | 85.6 | 85.4 | 78.2 | 28.30M | 6.96G |

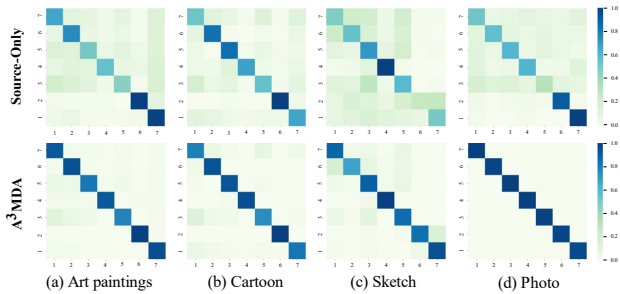

**Fig. 5.** Visualizations of similarity matrices on PACS.

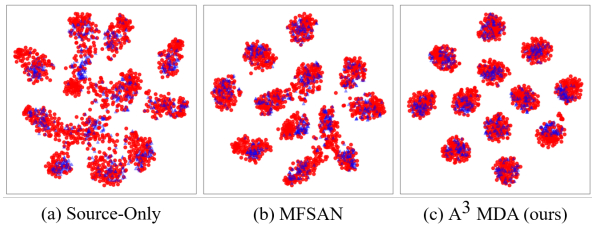

**Fig. 6.** The t-SNE visualizations of feature embeddings on '→I' task on ImageCLEF-DA (red: source; blue: target).

**Exploring Different Discrepancy Metrics.** The choice of discrepancy metrics is pivotal in discrepancy-based MDA methods. In $A^3$MDA, we compared traditional metrics—L2 distance [68] ('L2'), Wasserstein distance [69] ('WAS'), and CORAL [1] (Correlation Alignment)—with MMD on the Office-Home dataset. Results in Table X consistently demonstrate performance enhancements across these methods when employing our hardness-driven strategies, resulting in improvements of 5.9%, 4.0%, 4.6%, and 4.9%. These results affirm the potential of our approach in enhancing various conventional discrepancy-based MDA methods.

**Visualization of Similarity Matrix.** We present prototypical similarity matrices for four domains from the PACS dataset. As observed in Fig. 5, $A^3$MDA outperforms the Source-only baseline (top row) in capturing the underlying cluster-wise relationships, resulting in prominently reduced domain-specific noise in the matrices (bottom row). Notably, the noise reduction is most pronounced in the Photo domain, attributed to our cluster-level constraint that precisely performs intra-domain alignment in the pseudo-contrastive matrix, thereby enhancing its discriminative capability.

**Visualization of Feature Embeddings.** To validate the transferability of our model, we utilize t-SNE visualizations to depict the feature embeddings of different methods on the '→I' task of the ImageCLEF-DA dataset. As shown in Fig. 6, the target features learned by the Source-Only model exhibit a mismatch with the source domain. In contrast, our proposed method outperforms both Source-Only and MFSAN methods, evidenced by its generation of clusters with sharper boundaries. This demonstrates its superior transferability on the target domain while maintaining strong discrimination ability and performance.

**Visualization of Class Activation Mapping.** To enhance the interpretability, we present Grad-CAM results for two real-world images, depicted in Fig. 7. For alarm clocks (top row), our method sharply focuses on clock hands and partial circular edges, capturing crucial discriminative features. For bicycles (bottom row), our approach identifies and highlights both the front and back bicycles accurately, a detail overlooked by other methods. Overall, our approach effectively emphasizes features crucial for accurate classification.

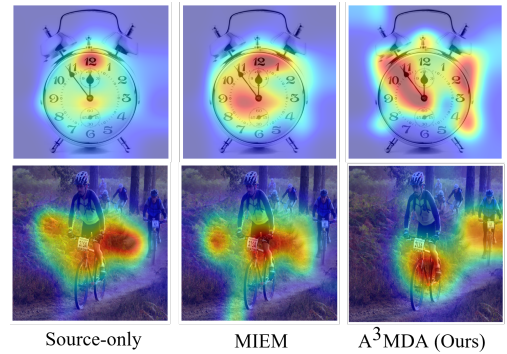

**Fig. 7.** Visualization of Grad-CAM on real-world images.

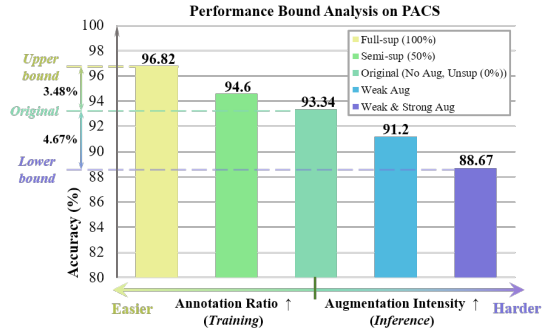

**Fig. 8.** Bound Analysis of performance on PACS.

**Performance Bound Analysis.** We conducted a bound analysis on PACS, which builds upper and lower bounds through varied approaches. For the upper bound, considering our model tackles an unsupervised setting where ground truth labels of the target domain are unseen during training, we progressively increased the proportion of visible labels to approximate peak performance. Specifically, we examined three annotation ratios: 0% (original unsupervised scenario), 50% (semi-supervised scenario, "Semi-sup"), and 100% (fully supervised scenario, "Full-sup", representing the upper bound). As for the lower bound, we applied diverse forms of data augmentation (aligned with those detailed in the methodology) to target domain samples during the inference stage to simulate extreme scenarios. Specifically, we considered three forms: no augmentation (original), weak augmentation ("Weak Aug"), and weak plus strong augmentation ("Weak & Strong Aug", representing the lower bound). As shown in Fig. 8, our method closely approaches the upper bound in the original unsupervised setting, with only a small gap of 3.48%. Besides, under strong noise perturbations, the performance only decreased by an average of 4.67%. This resilience can be attributed to our adaptive augmentation strategy, which effectively handles hard samples and ensures accurate predictions even under substantial perturbations. The above analysis underscores $A^3$MDA 's robust adaptability across varying levels of data availability and noise.

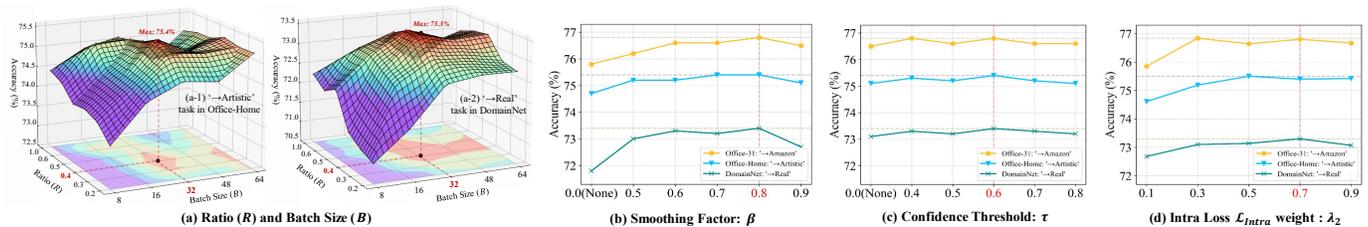

**Fig. 10.** (a) Joint sensitivity analysis of Batch Size ($B$) and Selection Ratio ($R$) on two tasks ('→Artistic' of Office-Home and '→Real' of DomainNet). Individual sensitivity analysis of (b) smoothing factor $\beta$, (c) pseudo-label threshold $\tau$ and (d) Intra loss weight $\lambda_2$ on three tasks ('→Amazon' of Office-31; '→Artistic' of Office-Home; '→Real' of DomainNet).

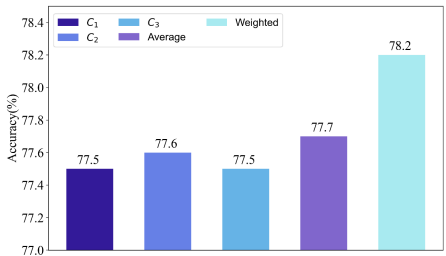

**Fig. 9.** Analysis of weighted prediction on Office-Home.

**Analysis of Prediction Approaches.** We also explore different prediction approaches used for inference. Apart from the weighted prediction $P_{w,i}^t$ used in A³MDA (referred to as 'Weighted'), we also consider the predictions $P_{m,i}^t$ generated by each domain-specific classifier $C_m(\cdot)$, as well as the average combination of these predictions $P_{a,i}^t = \frac{1}{M}\sum_{m=1}^M P_{m,i}^t$ (referred to as 'Average'). As depicted in Fig. 9, the weighted prediction performs the best as it dynamically determines the reliance of the target domain on different source domains based on their respective qualities. In contrast, solely relying on predictions from a domain-specific classifier $C_m(\cdot)$ or treating each domain equally without tailored processing with 'Average', would result in a biased or impartial approach that cannot achieve the benefits provided by dynamic weights $w_m$, leading to sub-optimal performance.

**Hyperparameters tuning strategies.** Our method employs a comprehensive three-level framework, necessitating the determination of multiple hyperparameters. Traditionally, a grid search approach across these hyperparameters could potentially enhance performance but is notably time-consuming and labor-intensive. Therefore, we adopted diverse tuning strategies to streamline the selection process:

**(1) Formula-based tuning for $\mathcal{L}_{Inter}$:** For $\lambda_1$, which governs the inter-domain alignment loss term $\mathcal{L}_{Inter}$, we applied a formula-based tuning strategy similar to DANN [2], using the adaptive formula $\lambda_1 = \frac{2}{exp(-\theta p)} - 1$ instead of a fixed value. According to [2], this tuning formula, along with the setting of $\theta = 10$, can effectively stabilize sensitivity during the early stages and has demonstrated success across several prominent MDA approaches (e.g., MFSAN [18]) for managing inter-domain alignment constraints.

**(2) Equalize weights of two MMD terms:** Initially, we introduced an additional hyperparameter $\eta$ to independently control the weighted clustered term in Eq. (7), transforming it to the form $\mathcal{L}_{Inter} = \mathcal{L}_{mmd} + \eta \cdot \mathcal{L}_{mmd}^{WC}$. Experimental findings indicated optimal model performance is observed when maintaining a balanced ratio of $\eta$ (within the range of $[0.8 - 1.2]$), with deviations outside this range resulting in slight performance decreases. Therefore, to simplify parameter tuning, we set $\eta$ to 1.0 (which can be seen as an omission).

**(3) Joint Tuning for $R$ and $B$:** The interaction between batch size ($B$) and hardness ratio ($R$) plays a crucial role in the construction of the Pseudo Contrastive Matrix. As shown in Fig. 10(a), we conducted a joint analysis by varying $B$ and $R$ across two tasks: '→Artistic' from Office-Home and '→Real' from DomainNet. As observed, excessively small batch sizes (e.g., $B \leq 16$) result in significant performance drops due to the limited number of samples, which hinders effective optimization of the overall loss function. On the other hand, while larger batch sizes (e.g., $B \geq 48$) reduce the intensity of competition among target samples in intra-domain alignment (due to the increased number of selected candidates to construct PCM), they still contribute positively to overall optimization due to the larger data volume processed per batch. For both tasks, the optimal performance was achieved with $B = 32$ and $R = 0.3$. This configuration strikes a balance by involving a sufficient number of samples in the competition, maintaining its focus, while also offering better computational efficiency compared to larger batch sizes.

**(4) Individual Tuning with appropriate tasks:** For the remaining hyperparameters, we perform individual tuning on three tasks: 'Amazon' from Office-31, 'Artistic' from Office-Home, and 'Real' from DomainNet. These tasks were selected based on three criteria: (1) moderate scale to ensure effective tuning; (2) sufficient task difficulty (e.g., the 'Amazon' in Office-31 is notably challenging); (3) representativeness, with characteristics shared across datasets (e.g., the 'Real' domain appears in both DomainNet and Office-Home).

According to Fig. 10(b), setting $\beta$ to 0 (i.e., replacing Smooth AHM with Basic AHM) may induce over-augmentation and performance decline due to the instability of instantaneous hardness measurements. Moreover, the commonly used value of 0.9 in standard EMA is not optimal, as an excessively large $\beta$ may overly smoothen hardness variations, complicating the adaptation of the model.

For $\tau$, unlike other pseudo-labeling approaches that are susceptible to varying threshold configurations, our method maintains a consistent accuracy within a narrow deviation of 0.5% across three datasets, as shown in Fig. 10(c). This stability is attributed to A³MDA's ability to gradually improve the quality of pseudo-labels through the proposed PCM.

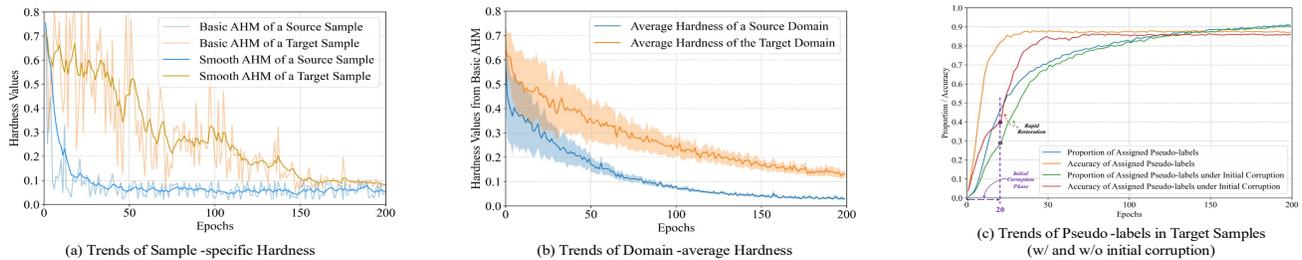

**Fig. 11.** (a) displays hardness values for a pair of samples from the target domain (Real) and one source domain (Clipart) in the '→Real' task of DomainNet. (b) presents the average hardness values in the above two domains. (c) shows the trends in accuracy and assignment rate of pseudo-labels (with and without corruption) in the target domain (Product) in '→Product' task of Office-Home.

For $\lambda_2$, which controls $\mathcal{L}_{Intra}$, we selected candidate values for $\lambda_2$ from the set $\{0.1, 0.3, 0.5, 0.7, 0.9\}$ and explore it on three specific tasks. As observed in Fig. 10(d), the overall best results consistently emerge around 0.7.

In conclusion, optimal performance is achieved with $B = 32$, $R = 0.4$, $\beta = 0.8$, $\tau = 0.6$, $\lambda_2 = 0.7$. Notably, these hyperparameters remained consistent across all seven datasets, delivering state-of-the-art performance in each. This indicates the strong generalizability of our chosen hyperparameter and the effectiveness of our optimization strategies, minimizing the necessity for task-specific adjustments.

**Analysis on Hardness Values.** Fig. 11(a) and Fig. 11(b) display trends in sample-specific and domain-average hardness. Early in training, the Basic AHM values for sample-specific hardness exhibit fluctuations due to the model's initial difficulty in handling strong perturbations, which results in unstable and low-confidence predictions. However, as training progresses, the model becomes well-fitted to both source and target samples, leading to more confident predictions and reduced fluctuations in Basic AHM values. In contrast, values from Smooth AHM are considerably smoother, providing a more stable measure of sample-specific hardness. For domain-average hardness, both the mean and standard deviation of Basic AHM values decrease for both domains as training advances. This reduction in perceived hardness values also correlates with increased data augmentation intensity, which reflects the enhanced generalization ability of the model.

**Analysis on Trends in Pseudo-labels.** Fig. 11(c) illustrates the trends of pseudo-labels for the target samples. Notably, we introduce an additional scenario where 50% of correct pseudo-labels are corrupted (randomly reassigned to other categories) during the first 20 epochs. In the uncorrupted setting, both the proportion of assigned pseudo-labels and their accuracy increase rapidly during initial training. In the corrupted setting, while the growth in accuracy is slightly slower, overall performance remains unaffected. This is because, in the early stages, the model focuses on coarse inter-domain alignment, which does not rely on high-quality pseudo-labels. As the assignment rate remains low, only a small number of corrupted pseudo-labels impact the cluster-level constraints in the current epoch, without significantly disrupting the overall training dynamics. As training progresses, the model refines its understanding of the target domain, and decision boundaries become more precise. In this phase, intra-domain shifts become the primary source of residual erroneous pseudo-labels. The adaptive intra-domain alignment then becomes more effective in correcting these labels. Ultimately, both settings demonstrate stable improvements in accuracy and assignment rates, leading to comparable performance.

## V. DISCUSSION

In this paper, we introduce $A^3MDA$, a novel framework for the MDA classification task. $A^3MDA$ addresses three key aspects: 1) the potential of data augmentation, 2) the importance of intra-domain alignment, and 3) the design of cluster-level constraints. To tackle these, we propose a series of adaptive hardness-driven strategies. For quantification, we progressively develop three Adaptive Hardness Measurements (AHMs): Basic, Smooth, and Comparative, each introducing new features at each stage of progression. As for utilization, instead of using measured hardness values for a single purpose, we adaptively incorporate hardness values in both augmentation and alignment scenarios through three actions, adjusting augmentation intensity to prevent over-augmentation and utilizing hardness values for cluster-level constraints.

$A^3MDA$ demonstrates broad practical applicability across three aspects. Firstly, in terms of dataset selection, we chose seven datasets that cover a wide range of real-world domains with diverse data distributions. $A^3MDA$ demonstrates superior performance across all benchmarks, showcasing its robustness in real-world applications. Secondly, our model effectively mitigates noise interference by adaptively incorporating various types of strong augmentation or noise during training. Thirdly, we also applied our hardness-driven framework to other feature extraction architectures and alignment constraints (e.g., divergence loss). We found that these adaptations also yield excellent performance and result in significant improvements compared to their baseline. These findings emphasize the broad applicability of our proposed hardness strategies in MDA classification tasks.

Despite its performance, $A^3MDA$ still has limitations. Firstly, the hardness-driven strategies require manual experimentation to match the most appropriate AHM for each scenario. We simplify this by thoroughly considering the characteristics introduced at each AHM, thus streamlining this matching process. Furthermore, due to its comprehensive nature, $A^3MDA$ deals with multiple constraints, complicating the hyperparameter selection. We expedite this process with various tuning strategies, enabling us to achieve a relatively optimal configuration across all datasets in a short timeframe. Finally, our framework focuses on image classification in the MDA setting, leaving out the adaptation of its strategy to cross-modal settings, which creates a challenging combination due to the compounded complexity of heterogeneous modalities and multiple source distributions. We envision our work as a paradigm for future research, where researchers can extend its elements to handle more challenging scenarios.

## V. Conclusion

In this paper, we present A³MDA, a novel hardness-driven strategy for Multi-source Domain Adaptation (MDA) that addresses three key aspects of conventional methods. A³MDA centers around adaptive quantification and utilization of hardness values. For quantification, we introduce three progressive Adaptive Hardness Measurements to effectively assess the hardness of source/target samples. For utilization, we adapt hardness values to various augmentation and alignment scenarios. Experiments showcase the potential of our adaptive hardness-driven strategy.